\def\year{2020}\relax
\documentclass[letterpaper]{article} 
\usepackage{aaai20}  
\usepackage{times}  
\usepackage{helvet} 
\usepackage{courier}  
\usepackage[hyphens]{url}  
\usepackage{graphicx} 
\usepackage{graphicx}  
\usepackage{amsmath,amssymb}
\usepackage{comment}
\usepackage{graphicx}
\usepackage{booktabs}
\usepackage{graphicx}
\usepackage{bm}
\usepackage{ifthen}
\usepackage{enumitem}
\usepackage{algorithm}
\usepackage{algorithmic}
\usepackage{tabularx,array,booktabs}
\usepackage{lipsum}  
\usepackage{tikz} 
\usetikzlibrary{arrows,decorations.pathmorphing,backgrounds,fit,positioning,shapes.symbols,chains}
\tikzset{box1/.style={draw=black, thick, rectangle, minimum height=1.4cm, minimum width=3cm }}

\newcommand{\Section}[1]{\section{#1}\label{sec:#1}}
\newcommand{\Subsection}[1]{\subsection{#1}\label{sec:#1}}
\newcommand{\Subsubsection}[1]{\subsubsection{#1}\label{sec:#1}}

\title{Bridging the gap between Markowitz planning and deep reinforcement learning}
\author{
	Eric Benhamou \textsuperscript{\rm 1,\rm 2}, David Saltiel \textsuperscript{\rm 1,\rm 3}, Sandrine Ungari \textsuperscript{\rm 4}, Abhishek Mukhopadhyay \textsuperscript{\rm 5} \\
	\textsuperscript{\rm 1} AI Square Connect, France, \texttt{\{eric.benhamou,david.saltiel\}@aisquareconnect.com} \\
	\textsuperscript{\rm 2}MILES, LAMSADE, Dauphine university, France, \texttt{eric.benhamou@lamsade.dauphine.fr} \\\
	\textsuperscript{\rm 3} LISIC, ULCO, France, \texttt{david.saltiel@univ-littoral.fr} \\
	\textsuperscript{\rm 4} Societe Generale, Cross Asset Quantitative Research, UK,\\ 
	\textsuperscript{\rm 5} Societe Generale, Cross Asset Quantitative Research, France,\\ 
	\texttt{\{sandrine.ungari,abhishek.mukhopadhyay\}@sgcib.com}
}

\begin{document}

\maketitle
\thispagestyle{plain}
\pagestyle{plain}

\begin{abstract}
While researchers in the asset management industry have mostly focused on techniques based on financial and risk planning techniques like Markowitz efficient frontier, minimum variance, maximum diversification or equal risk parity, in parallel, another community in machine learning has started working on reinforcement learning and more particularly deep reinforcement learning to solve other decision making problems for challenging task like autonomous driving, robot learning, and on a more conceptual side games solving like Go. 
This paper aims to bridge the gap between these two approaches by showing Deep Reinforcement Learning (DRL) techniques can shed new lights on portfolio allocation thanks to a more general optimization setting that casts portfolio allocation as an optimal control problem that is not just a one-step optimization, but rather a continuous control optimization with a delayed reward. The advantages are numerous: (i) DRL maps directly market conditions to actions by design and hence should adapt to changing environment, (ii) DRL does not rely on any traditional financial risk assumptions like that risk is represented by variance, (iii) DRL can incorporate additional data and be a multi inputs method as opposed to more traditional optimization methods. We present on an experiment some encouraging results using convolution networks.
\end{abstract}


\Section{Introduction}
In asset management, there is a gap between mainstream used methods and new machine learning techniques around reinforcement learning and in particular deep reinforcement learning. The former methods rely on financial risk optimization and solve the planning problem of the optimal portfolio as a single step optimization question. The latter do not make any assumptions about risk, do a more involving multi-steps optimization and solve complex and challenging tasks like autonomous driving \cite{Wang2018DeepRL}, learning advanced locomotion and manipulation skills from raw sensory inputs \cite{Levine_2015,Levine_2016,Schulman_2015,Schulman_2017,Lillicrap_2015} or on a more conceptual side for reaching supra human level in popular games like Atari \cite{mnih-atari-2013}, Go \cite{Silver_2016,silver2017mastering}, StarCraft II \cite{Vinyals_2019}, etc ... One of the reasons often put forward for this situation is that asset management researchers have mostly been trained with an econometric and financial mathematics background, while the deep reinforcement learning community has been mostly trained in computer science and robotics, leading to two distinctive research communities that do not interact much between each other. In this paper, we aim to present the various approaches to show similarities and differences to bridge the gap between these two approaches. Both methods can help solving the decision making problem of finding the optimal portfolio allocation weights.

\Subsection{Related works}
As this paper aims at bridging the gap between traditional asset management portfolio selection methods and deep reinforcement learning, there are too many works to be cited. 

On the traditional  methods side, the seminal work is \cite{markowitz1952portfolio} that has led to various extensions like minimum variance \cite{chopra1993effect,haugen1991efficient}, \cite{kritzman2014six}, maximum diversification \cite{choueifaty2008toward,choueifaty2012properties}, maximum decorrelation \cite{christoffersen2010is}, risk parity \cite{Maillard_2010,Roncalli_2016}. We will review these works in the section entitled \textit{Traditional methods}.

On the reinforcement learning side, the seminal book is \cite{SuttonBarto_2018}. The field of deep reinforcement learning is growing every day at an unprecedented pace, making the citation exercise complicated. But in terms of breakthroughs of deep reinforcement learning, one can cite the work around Atari games from raw pixel inputs \cite{mnih-atari-2013,Mnih_2015}, Go \cite{Silver_2016,silver2017mastering}, StarCraft II \cite{Vinyals_2019}, learning advanced locomotion and manipulation skills from raw sensory inputs \cite{Levine_2015,Levine_2016} \cite{Schulman_2015,Schulman_2016,Schulman_2017,Lillicrap_2015}, autonomous driving \cite{Wang2018DeepRL} and robot learning \cite{Gu_2017}.

On the application of deep reinforcement learning methods to portfolio allocations, there is already a growing interest as recent breakthroughs has put growing emphasis on this method. Hence, the field is growing very rapidly and survey like \cite{Fisher_2018} are already out dated. Driven initially mostly by applications to crypto currencies and Chinese financial markets \cite{Jiang_2016,Zhengyao_2017,Liang_2018,Yu_2019,Wang_2019}, the field is progressively taking off on other assets 
\cite{Kolm_2019,Liu_2020,Ye_2020,Li_2019,Xiong_2018}. More generally, DRL has recently been applied to other problems than portfolio allocation. For instance,  \cite{Deng_2016,zhang2019deep,huang2018financial,thate2020application,chakraborty2019capturing,nan2020sentiment,Wu_2020} tackle the problem of direct trading strategies \cite{bao2019multiagent} handles the one of multi agent trading while \cite{ning2018double} examine optimal execution.

\Section{Traditional methods}\label{sec:Traditional}
We are interested in finding an optimal portfolio which makes the planning problem quite different from standard planning problem where the aim is to plan a succession of tasks. Typical planning algorithms are variations around STRIPS \cite{Fikes_1971}, that starts by analysis ending goals and means, builds the corresponding graph and finds the optimal graph. Indeed we start from the goals to achieve and try to find means that can lead to them. New work like Graphplan as presented in \cite{Blum_1995} uses a novel planning graph, to reduce the amount of search needed, while hierarchical task network (HTN) planning leverages the classification to structure networks and hence reduce the number of graph searches. Other algorithms like search algorithm as $A^*$, $B^*$, weighted $A^*$ or for full graph search, branch and bound and its extensions, as well as evolutionary algorithms like particle swarm, CMA-ES are also used widely in AI planning etc.. However, when it comes to portfolio allocation, standard methods used by practitioners rely on more traditional financial risk reward optimization problems and follows rather the Markowitz approach as presented below.

\begin{figure}[!htbp]
\includegraphics[width= \linewidth] {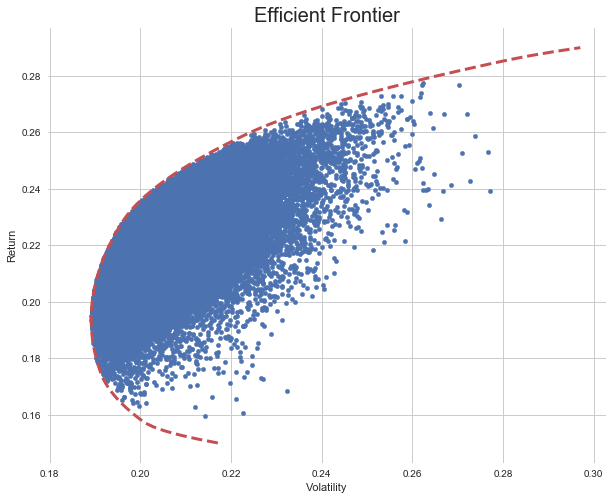}
\caption{Markowitz efficient frontier for the GAFA: returns taken from 2017 to end of 2019}\label{fig:Markowitz}
\end{figure}

\Subsection{Markowitz}
The intuition of Markowitz portfolio is to be able to compare various assets and assemble them taking into account both return and risk. Comparing just returns of some financial assets would be too naive. One has to take into account in her/his investment decision returns with associated risk. Risk is not an easy concept. in Modern Portfolio Theory (MPT), risk is represented by the variance of the asset returns. If we take various financial assets and display their returns and risk as in figure \ref{fig:Markowitz}, we can find an efficient frontier. Indeed there exists an efficient frontier represented by the red dot line.

Mathematically, if we denote by $w = (w_1, ..., w_l)$ the allocation weights with $1 \geq w_i \geq 0$ for $i=0 ... l$, summarized by $1 \geq w \geq 0 $, with the additional constraints that these weights sum to 1: $\sum_{i=1}^l w_i = 1$, we can see this portfolio allocation question as an optimization. \\
Let $\mu = ( \mu_1, ..., \mu_l)^T$ be the expected returns for our $l$ strategies and $\Sigma$ the matrix of variance covariances of the $l$ strategies' returns. Let $r_{min}$ be the minimum expected return. The Markowitz optimization problem to solve is to minimize the risk given a target of minimum expected return as follows:

\begin{eqnarray}
&\underset{w}{\text{Minimize}} & w^T \Sigma w      \label{eq:Markowitz}\\
&\text{subject to}& \mu^T w \geq r_{min}  ,  \sum_{i=1 \ldots l} w_i = 1, 1 \geq w \geq 0  \nonumber
\end{eqnarray}

It is solved by standard quadratic programming. Thanks to duality, there is an equivalent maximization with a given maximum risk $\sigma_{max}$ for wich the problem writes as follows:
\begin{eqnarray}
&\underset{w}{\text{Maximize}} & \mu^T w \label{eq:Markowitz2}\\
&\text{subject to} & w^T \Sigma w \leq \sigma_{max}  ,  \sum_{i=1 \ldots l} w_i = 1, 1 \geq w \geq 0  \nonumber
\end{eqnarray}

\Subsection{Minimum variance portfolio}
This seminal model has led to numerous extensions where the overall idea is to use a different optimization objective. As presented in \cite{chopra1993effect,haugen1991efficient}, \cite{kritzman2014six}, we can for instance be interested in just minimizing risk (as we are not so much interested in expected returns), which leads to the minimum variance portfolio given by the following optimization program:

\begin{eqnarray}
&\underset{w}{\text{Minimize}} & w^T \Sigma w  \label{eq:MinVariance}\\
&\text{subject to}& \sum_{i=1 \ldots l} w_i = 1, 1 \geq w \geq 0  \nonumber
\end{eqnarray}

\Subsubsection{Maximum diversification portfolio}
Denoting by $\sigma$ the volatilities of our $l$ strategies, whose values are the diagonal elements of the covariance matrix $\Sigma$: $\sigma = ( \Sigma_{i,i})_{i=1..l}$, 
we can shoot for maximum diversification with the diversification of a portfolio defined as follows: $D = \frac{w^{T}\sigma}{\sqrt{w^{T}\sum w}}$ . We then solve the following optimization program as presented in \cite{choueifaty2008toward,choueifaty2012properties} 

\begin{eqnarray}
&\underset{w}{\text{Maximize}} & \frac{w^{T}\sigma}{\sqrt{w^{T} \sum w}}  \label{eq:MaxDiversification}  \\
&\text{subject to}& \sum_{i=1 \ldots l} w_i = 1, 1 \geq w \geq 0 \nonumber
\end{eqnarray}

%
\noindent The concept of diversification is simply the ratio of the weighted average of volatilities divided by the portfolio volatility.

\Subsubsection{Maximum decorrelation portfolio}
Following \cite{christoffersen2010is} and denoting by $C$ the correlation matrix of the portfolio strategies, the maximum decorrelation portfolio is obtained by finding the weights that provide the maximum decorrelation or equivalently the minimum correlation as follows:

\begin{eqnarray}
&\underset{w}{\text{Minimize}} &  w^T C w   \label{eq:MaxDecorrel}\  \\
&\text{subject to}& \sum_{i=1 \ldots l} w_i = 1, 1 \geq w \geq 0  \nonumber
\end{eqnarray}

\Subsubsection{Risk parity portfolio}
Another approach following risk parity \cite{Maillard_2010,Roncalli_2016} is to aim for more parity in risk and solve the following optimization program 
\begin{eqnarray}
&\underset{w}{\text{Minimize}} & \frac{1}{2}  w^T \Sigma w  - \frac 1 n \sum_{i=1}^{l} \ln(w_i)  \label{eq:RiskParity}  \\
&\text{subject to}& \sum_{i=1 \ldots l} w_i = 1, 1 \geq w \geq 0 \nonumber
\end{eqnarray}

\noindent All these optimization techniques are the usual way to solve the planning question of getting the best portfolio allocation. We will see in the following section that there are many alternatives leveraging machine learning that remove cognitive bias of risk and are somehow more able to adapt to changing environment.

\Section{Reinforcement learning}
Previous financial methods treat the portfolio allocation planning question as a one-step optimization problem, with convex objective functions. There are multiple limitations to this approach: 
\begin{itemize}
\item they do not relate market conditions to portfolio allocation dynamically.
\item they do not take into account that the result of the portfolio allocation may potentially be evaluated much later.
\item they make a strong assumptions about risk.
\end{itemize}

What if we could cast this portfolio allocation planning question as a dynamic control problem where we have some market information and needs to decide at each time step the optimal portfolio allocation problem and evaluate the result with delayed reward? What if we could move from static portfolio allocation to optimal control territory where we can change our portfolio allocation dynamically when market conditions changes. Because the community of portfolio allocation is quite different from the one of reinforcement learning, this approach has been ignored for quite some time even though there is a growing interest for the use of reinforcement learning and deep reinforcement learning over the last few years. We will present here in greater details what deep reinforcement is in order to suggest more discussions and exchanges between these two communities.

Contrary to supervised learning, reinforcement learning do not try to predict future returns. It does not either try to learn the structure of the market implicitly. Reinforcement learning does more: it directly learns the optimal policy for the portfolio allocation in connection with the dynamically changing market conditions.

\Subsection{Deep Reinforcement Learning Intuition}
As it name stands for, Deep Reinforcement Learning (DRL) is the combination of Reinforcement Learning (RL) and Deep (D). The usage of deep learning is to represent the policy function in RL. In a nutshell, the setting for applying RL to portfolio management can be summarized as follows:
\begin{itemize}
    \item current knowledge of the financial markets is formalized via a state variable denoted by $s_t$.
    \item Our planning task which is to find an optimal portfolio allocation can be thought as taking an action $a_t$ on this market. This action is precisely the decision of the current portfolio allocation (also called portfolio weights).
    \item once we have decided the portfolio allocation, we observe the next state $s_{t+1}$.
    \item we use a reward to evaluate the performance of our actions. In our particular setting, we can compute this reward only at the the final time of our episode, making it quite special compared to standard reinforcement learning problem. We denote  this reward by $R_T$ where $T$ is the final time of our episode. This reward $R_T$ is in a sense similar to our objective function in traditional methods. A typical reward is the final portfolio net performance. It could be obviously other financial performance evaluation criteria like Sharpe, Sortino ratio, etc..
\end{itemize}

Following standard RL, we model our problem to solve with a Markov Decision Process (MDP) as in \cite{SuttonBarto_2018}. MDP assumes that the agent knows all the states of the environment and has all the information to make the optimal decision in every state. The Markov property implies in addition that knowing the current state is sufficient.
MDP assumes a 4-tuple $(\mathcal{S}, \mathcal{A}, \mathcal{P}, \mathcal{R})$ where $\mathcal{S}$ is the set of states, $\mathcal{A}$ is the set of actions, $\mathcal{P}$ is the state action to next state transition probability function $\mathcal{P} : \mathcal{S} \times \mathcal{A} \times \mathcal{S} \to \left[0, 1\right]$, and $\mathcal{R}$ is the immediate reward.  The goal of the agent is to learn a policy that maps states to the optimal action $\pi: \mathcal{S} \to \mathcal{A}$ 
and that maximizes the expected discounted reward $\mathbb{E}[\sum_{t=0}^{\infty }\gamma ^{t}R_t]$. 

The concept of using deep network is to represent the function that relates dynamically the states to the action called in RL the policy and denoted by $\vec a_t=\pi(s_t)$. This function is represented by deep network because of the universal approximation theorem that states that any function can be represented by a deep network provided we have enough layers and nodes. Compared to traditional methods that only solve a one step optimization, we are solving the following dynamic control optimization program:

\begin{eqnarray}
& \underset{\pi(.)}{\text{Maximize}}   & \mathbb{E}[R_T] \label{eq:DRL}\\
&\text{subject to}& a_t = \pi(s_t)\nonumber 
\end{eqnarray}

Note that we maximize the expected value of the cumulated reward  $\mathbb{E}[R_T]$ because we are operating in a stochastic environment. To make things simpler, let us assume that the cumulated reward is the final portfolio net performance. Let us write $P_t$ the price at time $t$ of our portfolio, and its return at time $t$: $r_t^P$ and the portfolio assets return vector  at time $t$: $\vec r_t$. The final net performance writes as  $P_T / P_0 - 1 = \prod_{t=1}^T (1+r_t^P) - 1$. The returns $r_t^P$ is a function of our planning action $a_t$ as follows: $(1+r_t^P) = 1 +
\langle \vec a_t , \vec r_t \rangle$ where $\langle \cdot,  \cdot \rangle$ is the standard inner product of two vectors. In addition if we recall that the policy is parametrized by some deep network parameters, $\theta$: $a_t = \pi_{\theta}(s_t)$, we can make our optimization problem slightly more detailed as follows:

\begin{eqnarray}
&\underset{\theta}{\text{Maximize}} & \mathbb{E}\left[ \prod_{t=1}^T \left( 1 +
\langle \vec a_t , \vec r_t \rangle \right) \right]   \label{eq:DRL2}\\
&\text{subject to}& a_t = \pi_{\theta}(s_t).\nonumber 
\end{eqnarray}

\noindent It is worth noticing that compared to previous traditional planning methods (optimization \ref{eq:Markowitz}, \ref{eq:MinVariance}, \ref{eq:MaxDiversification}, \ref{eq:MaxDecorrel} or \ref{eq:MaxDecorrel}), the underlying optimization problem in RL \ref{eq:DRL} and its rewritting in terms of deep network parameters $\theta$ as presented in  \ref{eq:DRL2} have many differences:

\begin{itemize}
\item First, we are trying to optimize a function $\pi$ and not simple weights $w_i$. Although this function at the end is represented by a deep neural network that has admittely also weights, this is conceptually very different as we are optimizing in the space of functions $\pi: \mathcal{S} \to \mathcal{A}$ , that is a much bigger space than simply $\mathbb{R}^l$.
\item Second, it is a multi time step optimization at it involves results from time $t=1$ to $t=T$, making it also more involving.
\end{itemize}

\Subsection{Partially Observable Markov Decision Process}
If there is in addition some noise in our data and we are not able to observe the full state, it is better to use Partially Observable Markov Decision Process (POMDP) as presented initially in \cite{Astrom_1969}. In POMDP, only a subset of the information of a given state is available. The partially-informed agent cannot behave optimally. He uses a window of past observations to replace states as in a traditional MDP.

Mathematically, POMDP is a generalization of MDP. 
POMPD adds two more variables in the tuple, $\mathcal{O}$ and $\mathcal{Z}$ where $\mathcal{O}$ is the set of observations and $\mathcal{Z}$ is the observation transition function $\mathcal{Z}: \mathcal{S} \times \mathcal{A} \times \mathcal{O} \to [0,1]$. 
At each time, the agent is asked to take an action $a_t \in \mathcal{A}$ in a particular environment state $s_t \in \mathcal{S}$, that is followed by the next state $s_{t+1}$ with $\mathcal{P}(s_{t+1}| s_t, a_t )$. The next state $s_{t+1}$ is not observed by the agent. It rather receives an observation $o_{t+1} \in \mathcal{O}$ on the state $s_{t+1}$ with probability $Z(o_{t+1}| s_{t+1}, a_t )$. 

From a practical standpoint, the general RL setting is modified by taking a pseudo state formed with a set of past observations $(o_{t-n}, o_{t-n-1}, \ldots, o_{t-1}, o_t)$. In practice to avoid large dimension and the curse of dimension, it is useful to reduce this set and take only a subset of these past observations with $j< n$ past observations, such that $0<i_1< \ldots < i_j$ and $i_k \in \mathbb{N}$ is an integer. The set $\delta_1 = (0,i_1, \ldots, i_j)$ is called the observation lags. In our experiment we typically use lag periods like (0, 1, 2, 3, 4, 20, 60) for daily data, where $(0,1,2,3,4)$ provides last week observation, $20$ is for the one-month ago observation (as there is approximately 20 business days in a month) and 60 the three-month ago observation.

\Subsection{Observations}
\Subsubsection{Regular observations}
There are two types of observations: regular and contextual information. Regular observations are data directly linked to the problem to solve. In the case of an asset management framework, regular observations are past prices observed over a lag period $\delta = (0<i_1< \ldots < i_j)$. To normalize data, we rather use past returns computed as  $r_t^k = \frac{p^k_t}{p^k_{t-1} } -1$ where $p^k_t$ is the price at time $t$ of the asset $k$. To give information about regime changes, our trading agent receives also empirical standard deviation computed over a sliding estimation window denoted by $d$ as follows $\sigma^k_t  = \sqrt{ \frac{1}{d} \sum_{u =t-d+1}^t \left( r_u - \mu \right)^2 }$, where the empirical mean $\mu$ is computed as $\mu = \frac{1}{d} \sum_{u =t-d+1}^t r_u$. Hence our regular observations is a three dimensional tensor $A_t = \left[ A^1_t, A^2_t\right]$
\begin{center}
with \,\, $A^1_t =  \left( \!
\begin{array}{c   }
r^1_{t-i_j} \,\,	... \,\, r^1_t \\
... \,\,... \,\, ...\\
r^m_{t-i_j} \,\,.... \,\, r^m_t
\end{array} \! \right)\! ,  \,\,
 A^2_t =  \left( \!
\begin{array}{c  }
\sigma^1_{t-i_j} 	\,\,	... \,\, \sigma^1_t\\
... \,\,... \,\, ...\\
\sigma^m_{t-i_j} \,\,.... \,\, \sigma^m_t
\end{array} \! \right)$
\end{center}
This setting with two layers (past returns and past volatilities) is quite different from the one presented in \cite{Jiang_2016,Zhengyao_2017,Liang_2018} that uses different layers representing closing, open high low prices. There are various remarks to be made. First, high low information does not make sense for portfolio strategies that are only evaluated daily, which is the case of all the funds. Secondly, open high low  prices tend to be highly correlated creating some noise in the inputs. Third, the concept of volatility is crucial to detect regime change and is surprisingly absent from these works as well as from other works like \cite{Yu_2019,Wang_2019,Liu_2020,Ye_2020,Li_2019,Xiong_2018}.

\Subsubsection{Context observation}\label{Context}
Contextual observations are additional information that provide intuition about current context. For our asset manager, they are other financial data not directly linked to its portfolio assumed to have some predictive power for portfolio assets. Contextual observations are stored in a 2D matrix denoted by $C_t$ with stacked past $p$ individual contextual observations. Among these observations, we have the maximum and minimum portfolio strategies return and the maximum portfolio strategies volatility. The latter information is like for regular observations motivated by the stylized fact that standard deviations are useful features to detect crisis. The contextual state writes as $
C^t =  \left( \!\!
\begin{array}{c  }
c^1_t 	\,\,	... \,\, c^1_{t-i_k}\\
... \,\,... \,\, ...\\
c^p_t 	\,\,.... \,\, c^p_{t-i_k}
\end{array} \!\!\! \right)$. The matrix nature of contextual states $C_t$ implies in particular that we will use 1D convolutions should we use convolutional layers. All in all, observations that are augmented observations, write as $O_t =[ A_t, C_t]$, with $A_t=[A^1_t, A^2_t]$ that will feed the two sub-networks of our global network.

\Subsection{Action}
In our deep reinforcement learning the augmented asset manager agent needs to decide at each period in which hedging strategy it invests. The augmented asset manager can invest in $l$ strategies that can be simple strategies or strategies that are also done by asset management agent. To cope with reality, the agent will only be able to act after one period. This is because asset managers have a one day turn around to change their position. We will see on experiments that this one day turnaround lag makes a big difference in results. As it has access to $l$ potential hedging strategies, the output is a $l$  dimension vector that provides how much it invest in each hedging strategy. For our deep network, this means that the last layer is a softmax layer to ensure that portfolio weights are between $0$ and $100\%$ and sum to $1$, denoted by $(p^1_t, ..., p^l_t)$. In addition, to include leverage, our deep network has a second output which is the overall leverage that is between 0 and a maximum leverage value (in our experiment 3), denoted by $lvg_t$. Hence the final allocation is given by $lvg_t \times (p^1_t, ..., p^l_t)$.

\Subsection{Reward}
In terms of reward, we are considering the net performance of our portfolio from $t_0$ to the last train date $t_T$ computed as follows: $\frac{P_{t_T}}{P_{t_0}}-1$.

\begin{figure}[!htbp]
\centering
\includegraphics[width= \linewidth]{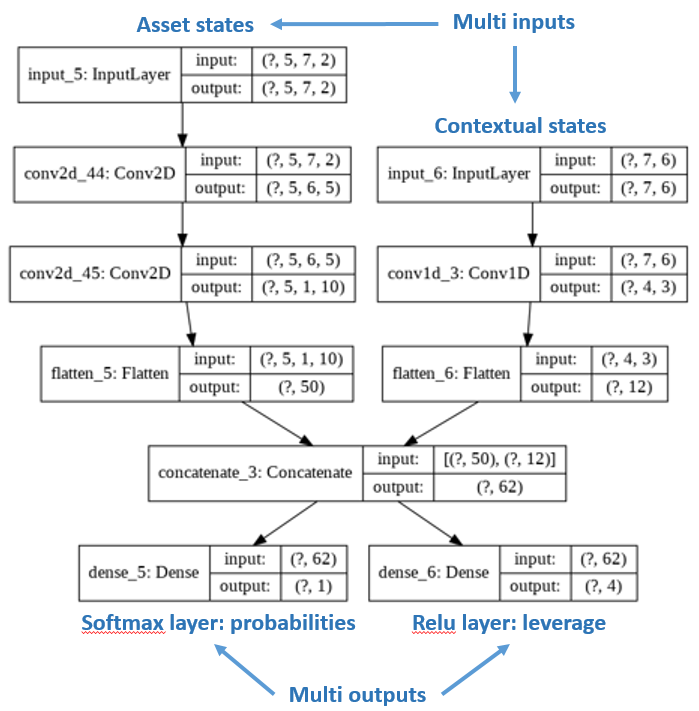}
\caption{network architecture obtained via tensorflow plotmodel function. Our network is very different from standard DRL networks that have single inputs and outputs. Contextual information introduces a second input while the leverage adds a second output}\label{fig:best_network}
\end{figure}

\Subsection{Multi inputs and outputs}
We display in figure \ref{fig:best_network} the architecture of our network. Because we feed our network with both data from the strategies to select but also contextual information, our network is a multiple inputs network.

Additionally, as we want from these inputs to provide not only percentage in the different hedging strategies (with a softmax activation of a dense layer) but also the overall leverage (with a dense layer with one single output neurons), we also have a multi outputs network. Additional hyperparameters that are used in the network as L2 regularization with a coefficient of 1e-8.

\Subsection{Convolution networks}
Because we want to extract some features implicitly with a limited set of parameters, and following \cite{Liang_2018}, we use convolution network that perform better than simple full connected layers. For our so called \textit{asset states} named like that because there are the part of the states that relates to the asset, we use two layers of convolutional network with 5 and 10 convolutions. These parameters are found to be efficient on our validation set. In contrast, for the contextual states part, we only use one layer of convolution networks with 3 convolutions. We flatten our two sub network in order to concatenate them into a single network.

\Subsection{Adversarial Policy Gradient}
To learn the parameters of our network depicted in \ref{fig:best_network}, we use a modified policy gradient algorithm called adversarial as we introduce noise in the data as suggested in \cite{Liang_2018}.. The idea of introducing noise in the data is to have some randomness in each training to make it more robust. This is somehow similar to drop out in deep networks where we randomly perturb the network by randomly removing some neurons to make it more robust and less prone to overfitting. Here, we are perturbing directly the data to create this stochasticity to make the network more robust.
A policy is a mapping from the observation space to the action space, $\pi:\mathcal{O}\rightarrow\mathcal{A}$. 
To achieve this, a policy is specified by a deep network with a set of parameters $\vec \theta$. The action is a vector function of the observation given the parameters: $\vec a_t = \pi_{\vec \theta}(\bm o_t)$.
The performance metric of $\pi_{\vec \theta}$ for time interval $[0,t]$ is defined as the corresponding total reward function of the interval $
	J_{[0,t]}(\pi_{\vec \theta}) = R\left( \vec o_1,\pi_{\vec \theta}(o_1),\cdots,
		\vec o_{t},\pi_{\vec \theta}(o_{t}),\vec o_{t+1} \right)
	\label{eq:policy_value}$.
After random initialization, the parameters are continuously updated along the gradient direction with a learning rate $\lambda$:
$\vec\theta \longrightarrow \vec\theta + \lambda\nabla_{\vec\theta}J_{[0,t]}(\pi_{\vec \theta})$. The gradient ascent optimization is done with standard Adam (short for Adaptive Moment Estimation) optimizer to have the benefit of adaptive gradient descent with root mean square propagation \cite{kingma2014method}. The whole process is summarized in algorithm \ref{alg1}.

\begin{algorithm}[!htbp]
    \caption{Adversarial Policy Gradient}
    \label{alg1}
\begin{algorithmic}[1]
    \STATE Input: initial policy parameters $\theta$, empty replay buffer $\mathcal{D}$
\REPEAT
    \STATE reset replay buffer
    \WHILE{not terminal}
        \STATE Observe observation $o$ and select action $a = \pi_{\theta}(o)$ with probability $p$ and random action with probability $1-p$, 
        \STATE Execute $a$ in the environment    
        \STATE Observe next observation $o'$, reward $r$, and done signal $d$ to indicate whether $o'$ is terminal
        \STATE apply noise to next observation $o'$
        \STATE store $(o,a,o')$ in replay buffer $\mathcal{D}$
        \IF{Terminal}
            \FOR{however many updates in $\mathcal{D}$}
                \STATE compute final reward $R$
            \ENDFOR
            \STATE update network parameter with Adam gradient ascent
                $\vec\theta \longrightarrow \vec\theta + \lambda\nabla_{\vec\theta}J_{[0,t]}(\pi_{\vec \theta})$
        \ENDIF
    \ENDWHILE
\UNTIL{convergence}
\end{algorithmic}
\end{algorithm}

In our gradient ascent, we use a learning rate of 0.01, an adversarial Gaussian noise with a standard deviation of 0.002. We do up to 500 maximum iterations with an early stop condition if on the train set, there is no improvement over the last 50 iterations.

\Section{Experiments}
\Subsection{Goal of the experiment}
We are interested in planing a hedging strategy for a risky asset. The experiment is using daily data from 01/05/2000 to 19/06/2020 for the MSCI and 4 SG-CIB proprietary systematic strategies. The risky asset is the MSCI world index whose daily data can be found on Bloomberg. We choose this index because it is a good proxy for a wide range of asset manager portfolios. The hedging strategies are 4 SG-CIB proprietary systematic strategies further described below. Training and testing are done following extending walk forward analysis as presented in \cite{Benhamou_202007} and \cite{benhamou2020time} with initial training from 2000 to end of 2006 and testing in a rolling 1 year period. Hence, there are 14 training and testing periods, with the different testing period corresponding to all the years from 2007 to 2020 and training done for period starting in 2000 and ending one day before the start of the testing period.

\Subsection{Data-set description}\label{Data-set}
Systematic strategies are similar to asset managers that invest in financial markets according to an adaptive, pre-defined trading rule. Here, we use 4 SG CIB proprietary 'hedging strategies', that tend to perform when stock markets are down:
\begin{itemize}
\item Directional hedges - react to small negative return in equities,
\item Gap risk hedges - perform well in sudden market crashes,
\item Proxy hedges - tend to perform in some market configurations, like for example when highly indebted stocks under-perform other stocks,
\item Duration hedges - invest in bond market, a classical diversifier to equity risk in finance. 
\end{itemize}

The underlying financial instruments vary from put options, listed futures, single stocks, to government bonds. Some of those strategies are akin to an insurance contract and bear a negative cost over the long run. The challenge consists in balancing cost versus benefits.

In practice, asset managers have to decide how much of these hedging strategies are needed on top of an existing portfolio to achieve a better risk reward. The decision making process is often based on contextual information, such as the economic and geopolitical environment, the level of risk aversion among investors and other correlation regimes. The contextual information is modeled by a large range of features :

\begin{itemize}
    \item the level of risk aversion in financial markets, or market sentiment, measured as an indicator varying between 0 for maximum risk aversion and 1 for maximum risk appetite,
    \item the bond equity historical correlation, a classical ex-post measure of the diversification benefits of a duration hedge, measured on a 1 month, 3 month and 1 year rolling window,
    \item The credit spreads of global corporate - investment grade, high yield, in Europe and in the US - known to be an early indicator of potential economic tensions, 
    \item The equity implied volatility, a measure if the 'fear factor' in financial market,
    \item The spread between the yield of Italian government bonds and the German government bond, a measure of potential tensions in the European Union,
    \item The US Treasury slope, a classical early indicator for US recession,
    \item And some more financial variables, often used as a gauge for global trade and activity: the dollar, the level of rates in the US, the estimated earnings per shares (EPS).
    
\end{itemize}
A cross validation step selects the most relevant features. In the present case, the first three features are selected. The rebalancing of strategies in the portfolio comes with transaction costs, that can be quite high since hedges use options. Transactions costs are like frictions in physical systems. They are taken into account dynamically to penalise solutions with a high turnover rate.

\Subsection{Evaluation metrics}
Asset managers use a wide range of metrics to evaluate the success of their investment decision. For a thorough review of those metrics, see for example \cite{Cogneau_2009}. The metrics we are interested in for our hedging problem are listed below:
\begin{itemize}
\item annualized return defined as the average annualized compounded return,
\item annualized daily based Sharpe ratio defined as the ratio of the annualized return over the annualized daily based volatility $\mu / \sigma$,
\item Sortino ratio computed as the ratio of the annualized return overt the downside standard deviation,
\item maximum drawdown (max DD) computed as the maximum of all daily drawdowns. The daily drawdown is computed as the ratio of the difference between the running maximum of the portfolio value defined as $RM_T = \max_{t=0..T}(P_t)$ and the portfolio value over the running maximum of the portfolio value. Hence the drawdon at time $T$ is given by $DD_T = (RM_T - P_T) / RM_T$ while the maximum drawdown $MDD_T = \max_{t=0..T}(DD_t)$. It is the maximum loss in return that an investor will incur if she/he invested at the worst time (at peak).
\end{itemize}

\Subsection{Results and discussion}\label{sec:Results}
Overall, the DRL approach achieves much better results than traditional methods as shown in table \ref{tab:Model comparison}, except for the maximum drawdown (max DD). Because time horizon is important in the comparison we provide risk measures for the last 2 and 5 years to emphasize that the DRL approach seems more robust than traditional portfolio allocation methods.

\begin{table}[!htbp]
  \centering
  \caption{Models comparison over 2 and 5 years}\label{tab:Model comparison}
\resizebox{\linewidth} {!} {
    \begin{tabular}{|l|rrrr|}
    \toprule
 &        & 2 Years &   &\\
    \toprule   
        & \multicolumn{1}{l}{return } & \multicolumn{1}{l}{Sortino} & \multicolumn{1}{l}{Sharpe} & \multicolumn{1}{l|}{max DD}\\
    Risky asset & 8.27\% &           0.39  &           0.36  & -        0.34  \\
    DRL    & \textbf{20.64\%} & \textbf{0.94 } & \textbf{0.96 } & -        0.27  \\
    Markowitz & -0.25\% & -        0.01  & -        0.01  & -        0.43  \\
    MinVariance & -0.22\% & -        0.01  & -        0.01  & -        0.43  \\
    MaxDiversification & 0.24\% &           0.01  &           0.01  & -        0.43  \\
    MaxDecorrel & 14.42\% &           0.65  &           0.63  & -        0.21  \\
    RiskParity & 14.17\% &           0.73  &           0.72  & \textbf{-0.19 } \\
    \toprule
     &        & 5 Years &   &\\
      \toprule   
        & \multicolumn{1}{l}{return } & \multicolumn{1}{l}{Sortino} & \multicolumn{1}{l}{Sharpe} & \multicolumn{1}{l|}{max DD}\\
    Risky asset & 9.16\% &           0.57  &           0.54  & -        0.34  \\
    DRL    & \textbf{16.95\%} & \textbf{1.00 } & \textbf{1.02 } & -        0.27  \\
    Markowitz & 1.48\% &           0.07  &           0.06  & -        0.43  \\
    MinVariance & 1.56\% &           0.08  &           0.06  & -        0.43  \\
    MaxDiversification & 1.77\% &           0.08  &           0.07  & -        0.43  \\
    MaxDecorrel & 7.65\% &           0.44  &           0.39  & -        0.21  \\
    RiskParity & 7.46\% &           0.48  &           0.43  & \textbf{-0.19 } \\
    \bottomrule
    \end{tabular}
}
\end{table}%

\begin{figure}[!htbp]
\centering
\includegraphics[width= \linewidth]{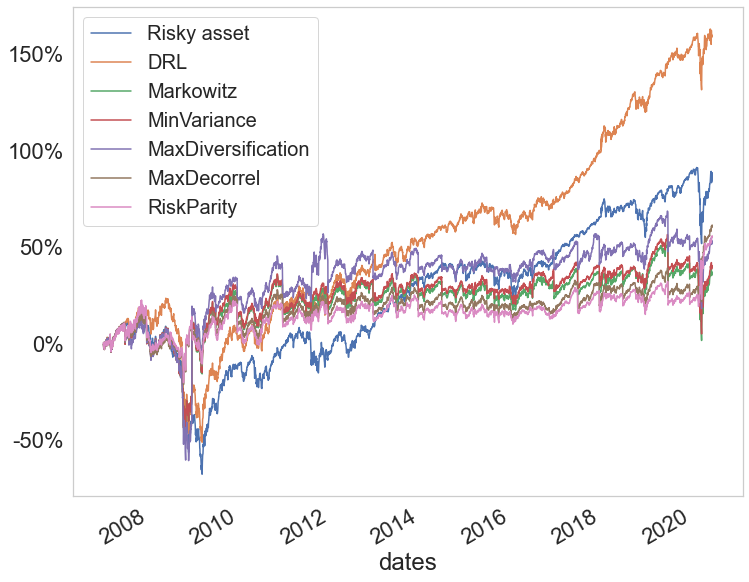}
\caption{performance of all models}\label{fig:all_models}
\end{figure}

When plotting performance results from 2007 to 2020 as shown in figure \ref{fig:all_models}, we see that DRL model is able to deviate upward from the risky asset continuously, indicating a steady performance. In contrast, other financial models are not able to keep their marginal over-performance over time with respect to the risky asset and end slightly below the risky asset.

\begin{figure}[!htbp]
\centering
\includegraphics[width= 0.495 \linewidth]{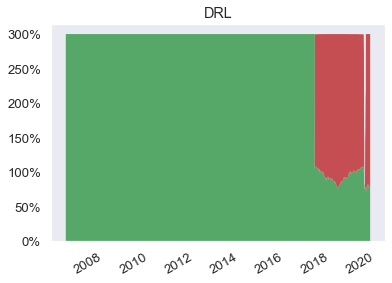}
\includegraphics[width= 0.495 \linewidth]{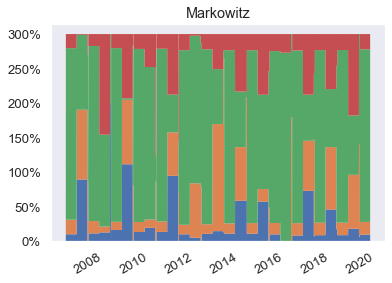}
\includegraphics[width= 0.495 \linewidth]{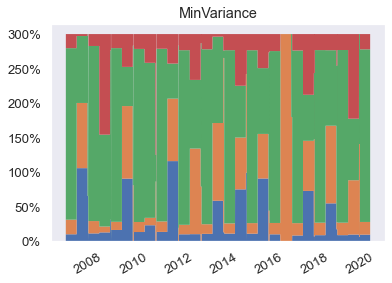}
\includegraphics[width= 0.495 \linewidth]{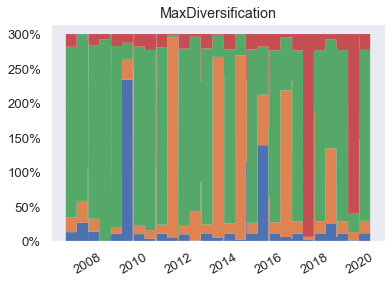}
\includegraphics[width= 0.495 \linewidth]{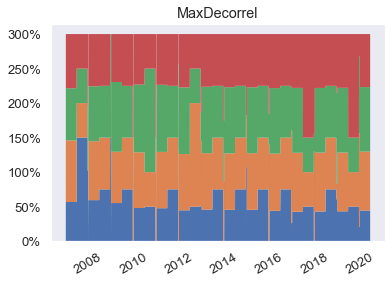}
\includegraphics[width= 0.495 \linewidth]{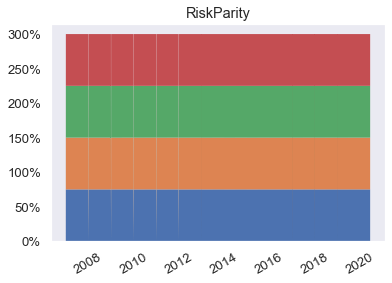} \\
\includegraphics[width= 0.4 \linewidth]{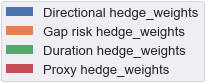}
\caption{weights for all models}\label{fig:weights}
\end{figure}

\Subsection{Allocation chosen by models}
The reason of the stronger performance of DRL comes from the way it chooses its allocation. Contrarily to standard financial methods that play the diversification as shown in figure \ref{fig:weights}, DRL aims at choosing a single hedging strategy most of the time and at changing it dynamically, should the financial market conditions change. In a sense, DRL is doing some cherry picking by selecting what it thinks is the best hedging strategy.

In contrast, traditional models like Markowitz, minimum variance, maximum diversification, maximum decorrelation and risk parity provides non null weights for all our hedging strategies and do not do cherry picking at all. They are neither able to change the leverage used in the portfolio as opposed to DRL model.

\begin{figure}[!htbp]
\centering
\includegraphics[width= 0.7\linewidth]{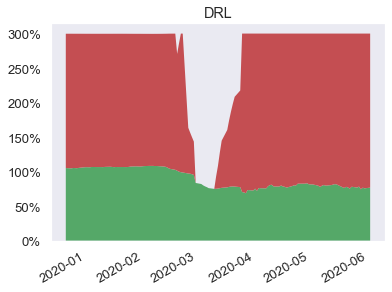}
\caption{disallocation of DRL model}\label{fig:drl_disallocation}
\end{figure}

\Subsection{Adaptation to the Covid Crisis}
The DRL model can change its portfolio allocation should the market conditions change. This is the case from 2018 onwards, with a short deleveraging window emphasized by the small blank disruption during the Covid crisis as shown in figure \ref{fig:drl_disallocation}. We observe in this figure where we have zoomed over year 2020, that the DRL model is able to reduce leverage from 300 \% to 200 \% during the Covid crisis (end of February 2020 to start of April 2020). This is a unique feature of our DRL model compared to traditional financial planning models that do not take leverage into account and keeps a leverage of 300 \% regardless of market conditions.

\Subsection{Benefits of DRL}
As illustrated by the experiment, the advantages of DRL are numerous: (i) DRL maps directly market conditions to actions by design and hence should adapt to changing environment, (ii) DRL does not rely on any traditional financial risk assumptions, (iii) DRL can incorporate additional data and be a multi inputs method as opposed to more traditional optimization methods.

\Subsection{Future work}
As nice as this work is, there is room for improvement as we have only tested a few scenarios and only a limited set of hyper-parameters for our convolutional networks. We should do more intensive testing to confirm that DRL is able to better adapt to changing financial environment. We should also investigate the impact of more layers and  other design choice in our network.

\Section{Conclusion}
In this paper, we discuss how a traditional portfolio allocation problem can be reformulated as a DRL  problem, trying to bridge the gaps between the two approaches. We see that the DRL approach enables us to select fewer strategies, improving the overall results as opposed to traditional methods that are built on the concept of diversification. We also stress that DRL can better adapt to changing market conditions and is able to incorporate more information to make decision. 

\subsection{Acknowledgments}
We would like to thank Beatrice Guez and Marc Pantic for meaningful remarks. The views contained in this document are those of the authors and do not necessarily reflect the ones of SG CIB.

\bibliography{main}
\bibliographystyle{aaai}
\end{document}